\documentclass[fleqn,11pt]{wlscirep}
\usepackage[utf8]{inputenc}
\usepackage[T1]{fontenc}
\usepackage{graphicx}
\usepackage{cite}
\usepackage{amsmath,amssymb,amsfonts}
\usepackage{algorithmic}
\usepackage{textcomp}
\usepackage{url}
\usepackage[colorinlistoftodos]{todonotes}
\usepackage{bm}%
\usepackage{stmaryrd} %
\usepackage{multirow}
\usepackage[flushleft]{threeparttable}
\usepackage{mwe}
\usepackage{xr}
\usepackage{longtable}%
\usepackage[ruled]{algorithm}
\usepackage{bm}
\usepackage{multicol}
\usepackage{arydshln}
\usepackage{wrapfig,lipsum,booktabs}
\usepackage{subcaption}
\usepackage{tabularx, makecell}
\usepackage{pbox}
\usepackage{caption}

\usepackage{mathrsfs}

\begin{document}

\title{Online-to-Offline Advertisements as Field Experiments}

\author[1*]{Akira Matsui}
\affil[1]{Department of Computer Science, University of Southern California, Los Angeles, CA, USA}

\author[2]{Daisuke Moriwaki}
\affil[2]{AI Lab, CyberAgent, Inc., Shibuya, Tokyo, Japan.}

\affil[*]{Corresponding author: amatsui@usc.edu}
\begin{abstract}

Online advertisements have become one of today's most widely used tools for enhancing businesses partly because of their compatibility with A/B testing. A/B testing allows sellers to find effective advertisement strategies such as ad creatives or segmentations. Even though several studies propose a technique to maximize the effect of an advertisement, there is insufficient comprehension of the customers' offline shopping behavior invited by the online advertisements. Herein, we study the difference in offline behavior between customers who received online advertisements and regular customers (i.e., the customers visits the target shop voluntary), and the duration of this difference. We analyzed approximately three thousand users' offline behavior with their 23.5 million location records through 31 A/B testings. We first demonstrate the externality that customers with advertisements traverse larger areas than those without advertisements, and this spatial difference lasts several days after their shopping day. We then find a long-run effect of this externality of advertising that a certain portion of the customers invited to the offline shops revisit these shops. Finally, based on this revisit effect findings, we utilize a causal machine learning model to propose a marketing strategy to maximize the revisit ratio. Our results suggest that advertisements draw customers who have different behavior traits from regular customers. This study's findings demonstrate that a simple analysis may underrate the effects of advertisements on businesses, and an analysis considering externality can attract potentially valuable customers.
\end{abstract}

\maketitle

\section{Introduction}
\label{sec:intro}
Online advertisements play an integral role in today's World Wide Web. Regardless of whether they are for online or offline shopping, the center of interest in ad technology developments thus far is to increase the sales of the targeted items or shops. Therefore, even post hoc analysis of the advertisements mainly focuses on these outcomes. However, to propose an effective marketing strategy, we need to disentangle what type of the customers advertisements invite to the target shop. The customers invited by the advertisement may show different shopping patterns from customers who visit the shop voluntarily. If these differences exist, an appropriate advertisement strategy for each group is not the same. Studying customer behavior during their shopping will advance understanding of what customers are induced by the advertisements and the additional marketing strategy that should be adopted to enhance the business.

Recent studies, for instance, have investigated customers' offline shopping trajectories, who have received advertisements. The customers who received advertisements increase in-store traffic, resulting in unplanned purchases~\cite{hui2013the,hui2013deconstructing}. The field experiment in a shopping mall suggests a similar phenomenon in which location-based advertisements change customers' shopping behavior trajectories\cite{ghose2019mobile}.
These studies imply that just focusing on the traditional simple measurement may neglect the behavioral differences between regular customers and the customers who are invited to the shops by the advertisement.

Although the above literature provides insightful findings on customers' shopping behavior, several vital issues remain unexplored. First, they only cover a limited area for shopping, such as in a grocery store~\cite{hui2013the, hui2013deconstructing} or shopping mall~\cite{ghose2019mobile}. Additionally, they only focus on the trajectories of the target shop visit day, and not much is known about the long term trajectory of the customers who received advertisements. In other words, few studies have focused on the entire picture of the long-run customers' trajectory. However, such comprehension of customers' offline behavior is essential to maximize the effectiveness of brick-and-mortar advertising. For example, suppose a customer visits several stores when they visit the advertised shop. In that case, additional advertisements for those other stores would have a better conversion rate. In addition, analyzing long-term offline trajectories enables us to understand the type of customers who are likely to revisit the store (i.e., those who will become loyal customers). This long term analysis will enable us to maximize repeat customers through advertisements.

To fill the gaps mentioned above, we analyzed the long-term customer trajectory data from the mobile devices that consist of 4 months of customers' location data. We first map the experimental data for Online to Offline (O2O) advertisements and the customers' location data. Thereafter, we address the following three major research questions:

\begin{description}

  \item[RQ1 Spatial trajectory:] Do the customers invited by the advertisement show different shopping trajectories than regular customers on the target shop visit day?
  \item[RQ2 Persistence of trajectory difference:] How long does that trajectory difference last?
  \item[RQ3 Revisiting:] How likely customers will return to the same shop after the visit, and what factors are associated with that revisiting behavior?

\end{description}
With the answers to the third research question, we will utilize our findings to propose an effective advertising strategy that maximizes the number of revisit customers. To this end, we use an uplift model, a framework of causal machine learning, to propose an advertisement campaign strategy aiming to maximize the revisiting effect.

This paper contributes to the current state-of-the-art in three ways. First, we study the trajectory differences between the customers invited by the advertisement and regular customers who voluntarily visit the shop on a large scale. With the long-term user trajectory data, we also study the duration of the trajectory differences between the two groups. Our holistic analysis implies that the existing studies omit unignorable behavioral differences between the two groups. We also discovered that a certain fraction of the users invited to the physical shop revisit the same shop. Utilizing this discovery, we propose a prediction model that maximizes the revisiting effect of the advertisement.

\section{Study Design \& Data}

In this section, we report our study design and data. We first discuss our strategy to answer the three research questions introduced in Sec.~\ref{sec:intro}. Thereafter, we explain our dataset, which consists of the A/B testings for the advertisements. We also introduce a glossary of terms in Table~\ref{table:glossary} to make our deliberations in this paper consistent and concise.

\subsection{Study Design to Answer the RQs}\label{sec:study_design}
Our three research questions ask if the difference in the motivation for visiting the shops reflect the trajectory differences. To answer these questions, we compose the A/B testing data for O2O advertisements. 

As a regular A/B testing, the users in the experiment are randomly split into two groups to study the effectiveness of the O2O advertisements. One group revcieves the advertisements (Treatment), whereas the other group does not distribute the advertisements (Control). The primary interest of these A/B testing is the target shops that the online advertisement causes customers to visit. To understand this point, the location data is collected from the users' mobile devices to check whether the users actually visit the target shop.

\begin{figure}[t]
    \centering
  \includegraphics[width=0.4\columnwidth, height=0.25\linewidth]{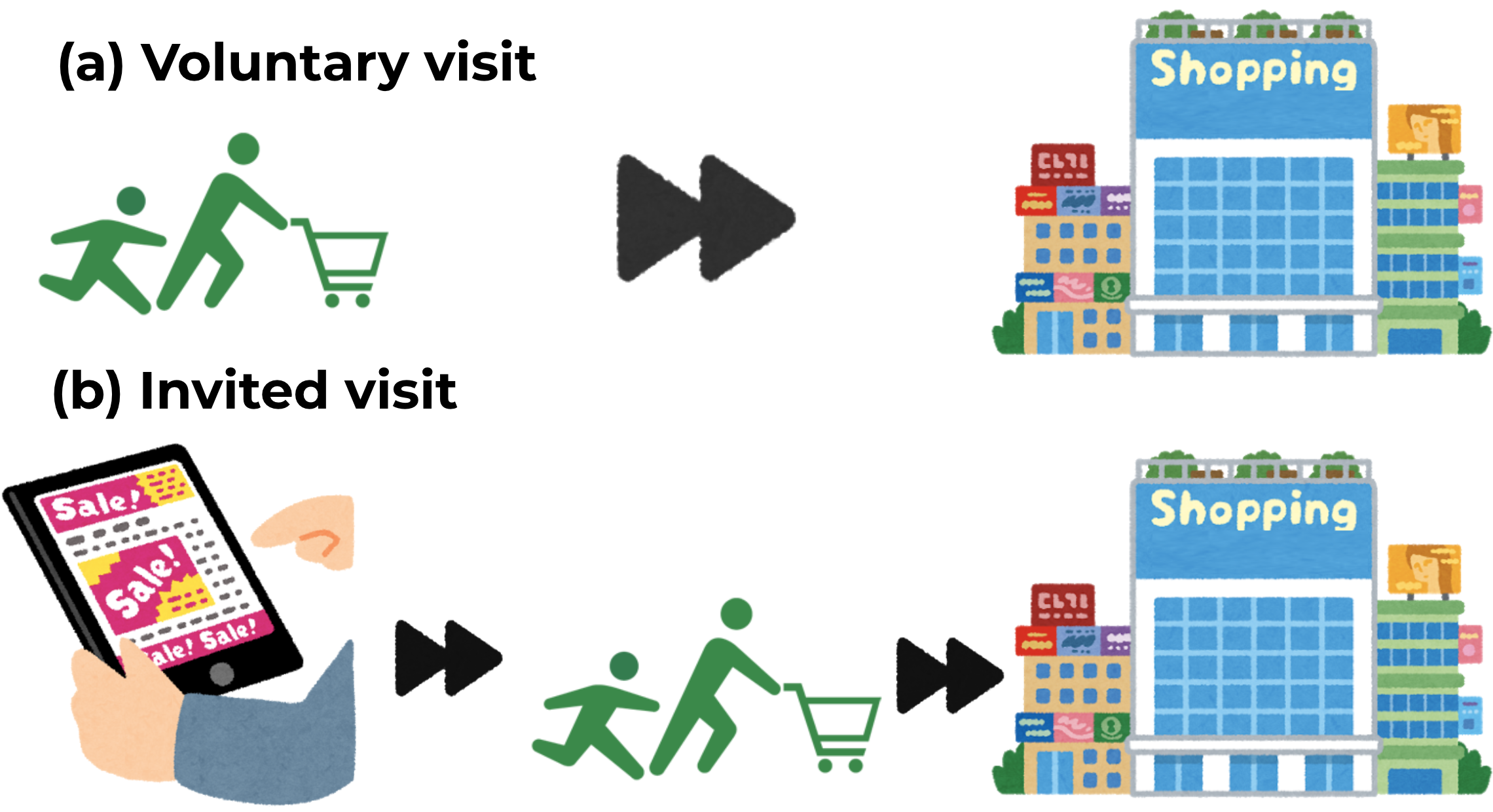}
  \caption{Schematic of the voluntary and invited visit} 
  \label{fig:schematic}
    \begin{minipage}{\columnwidth}%
    {\small {\it Note}: Schematic of our research design. We compare the two groups that have different motivations to visit the target shop. The users in \textbf{(a) Voluntary visit} visit the target shop without seeing any O2O advertisement. The users in \textbf{(a)} are supposed to have an intrinsic motivation for shopping. On the other hand,  \textbf{(b) Invited visit} group is the users that receive the O2O advertisements, and hence they are supposed to have an extrinsic motivation for shopping. In this study, we study if this motivational difference reflect the trajectory difference.}
  \end{minipage}%
\end{figure}

In this study, we investigate the experimental data from a different point of view, summarized in the schematic example in Figure~\ref{fig:schematic}. We leverage the situation in which the A/B testing for O2O advertisements causes motivational distinctions. The experimental data contains the users of the two groups, and this difference corresponds to the motivation to visit the shop. The control group users are supposed to visit the target shop voluntarily because any O2O advertisement is not distributed to this group. In contrast, the treatment group can have an extrinsic motivation provided by the distributed advertisement. To understand if such motivational difference reflects their offline trajectories, we compare the trajectory between the voluntary customers (Control) and those of the invited customers (Treatment). Table~\ref{table:glossary} summarizes the terms that we introduced in this section.

\begin{table}[!htbp] \footnotesize
\caption{Glossary of Terms}
\label{table:glossary} 
  \centering
    \begin{tabularx}{\columnwidth}{lX}
    \textbf{Terminology} & \textbf{Description} \\
    \toprule
O2O advertisement & Offline-to-Online advertisement for physical shops\\
Target shop & A physical shop advertised by a O2O advertisement. \\
Target shop visit day & The day when a customer visited the target shop. In this study,  we focus on the first target shop visit day for RQ1 and RQ2 and study the revisit behavior for RQ3.\\
Invited customers & The customers that the advertisements invited to the target shops (i.e., the treatment group) \\
Voluntary visit customers & The customers who visit the target shop voluntary. (i.e., the control group) \\
    \bottomrule
    \end{tabularx}
    
\begin{minipage}{\columnwidth}%
    {\footnotesize {\it Note}: This glossary reports the definition of the terms used in this study. To describe our study consistently and concisely, we define our user groups and subjects in a simple wordings.}
  \end{minipage}%
\end{table}

\subsection{Volume of the Audiences and the Resulting Sales}~\label{sec:audience}
We conduct our study based on the data generated from the A/B testing of the O2O advertisements conducted by CyberAgent, a Japan-based major advertising agency. They conduct nation-wide advertisement campaigns for offline shopping, aiming at inviting potential customers to a physical shop through online advertisements. Their campaign distributes online advertisements based on customer behavior.

As typical online advertisements, the advertisement campaigns use a demand-side platform (DSP) to deliver their advertisement creatives to potential customers' mobile devices through a real-time bidding (RTB) system. During the campaigns, users' visiting histories are used to count the number of visits to target stores. To understand the effectiveness of each campaign, we randomly split the user pool into two groups as follows: treatment and control. Only the users in the treatment group have a chance to receive the advertisements, whereas those in the control group do not receive any treatment.

We study the multiple campaigns for multiple businesses. They cover the various categories from electronic stores to super markets~\footnote{The number of campaigns (i.e., experiments) with the category of the target shop; Variety store: 15; Sport good store: 6; Electronic store: 6; Housing-related store: 2; Super market: 1; Store in other category: 2}. As we introduced in Sec.~\ref{sec:study_design}, this study focuses on the customers who visited a target shop, resulting in the number of customers in the control group being 1,440 and the number of customers in the treatment group is 1,515.

\subsection{Offline Behavior Trajectory Data}~\label{sec:offline_tra_data}

In addition to the visiting history to the target shop, we are interested in their trajectories near the target shop visit day. We collect the customer location data from their mobile devices and label them with the actual commercial place data. We trace the customers' shopping trajectories, matching the location data from the customer with the commercial place data set. We then only label ``visited'' if a customer stayed for more than 10 min within a 20 m radius of a location registered in the shop dataset. In addition, we label the visited location with the three commercial categories as follows: shopping, food, and service. Consequently, our dataset contains 23,549,902 location histories for 2,955 users from January 1, 2020 to April 1, 2020. For the additional analysis, we also study the post-experiment user location data from April 1, 2020 to July 31, 2020 for the revisit analysis to be described in the subsequent section.

\subsection{Offline Behavior Features for the Uplift Model}\label{sec:upliftmodel_feature}

We also compose the users' behavior features built on the user offline trajectory data for uplift modeling to be detailed at Sec.~\ref{method:uplift}. To prevent the data leakage, we do not use any data for the analysis for \textbf{RQ1-3} described in Sec.~\ref{sec:audience} and \ref{sec:offline_tra_data}. The demographic features are predicted values for each demographic characteristics that are provided by CyberAgent, Inc. The details of the prediction models are not provided due to the business reasons, but we would like here to mention that it is well know that the offline trajectories of the users can predict their demographic information~\cite{raviPredictingConsumerLevel2018c,zhong2015you,wang2017inferring,hinds2018demographic,bao2016geo}. Note that this estimation results are used only for the uplift and \textbf{not for our main research questions (RQ1-3)}. In addition, we use 96 shopping categories of the shops that the users visited which covers shopping categories from Fishing Shop to Bakery. Finally, we estimate the users' home location and calculate the distance from their home to the target shops. We only use the data before the period of the experiments to prevent data leaking.

\section{methods}
In this section, we describe the methods to answer the 
aforementioned three research questions. For \textbf{RQ1}, we discuss our spatial trajectory analysis at Sec.~\ref{sec:sptr}. Then we explain our panel regression model to study \textbf{RQ2} at Sec.~\ref{method:travel_distance}. Lastly, for \textbf{RQ3}, we propose the method to calculate the revisit effect of the O2O advertisements at Sec.~\ref{method:re-visit}, and the causal machine learning model to maximize the lift effect of the revisit at Sec.~\ref{method:uplift}.

\subsection{RQ1: Spatial Trajectory Analysis}\label{sec:sptr}
With the compiled shopping trajectory data, we studied whether the invited customers show different shopping trajectories compared with the voluntary customers. To this end, we model customers' spatial trajectory. We first align the user location data such that we could analyze the customer behavior on the same coordinate and run a regression model to study the spatial features of each group.

\subsubsection{Aligning spatial shopping trajectories from the different places: }\label{method:coordination}
As discussed in the previous sections, the experiments target multiple offline shops and are on a nationwide scale. This makes it impossible to map the customers' movement in one actual map as a typical geological analysis does. In other words, the raw trajectory data contains the user location history in different places across the nation, and hence, they do not allow us to summarize their trajectories as a single model. Therefore, we align the customer trajectory data into the same coordinate. 

We refer to the location of the target shop for this alignment. For each customer travel point $(u_i, v_i)$, we align it taking longitude and latitude differences as follows: 

\begin{eqnarray}\label{eq:latlon}
(u_i, v_i) = (u^{raw}_{i} - u^{ref}_{i}, v^{raw}_{i} - v^{ref}_{i})
\end{eqnarray}

where $u^{l}_{i}$ and $v^{l}_{i}$ denote the latitude and longitude of the $l$, $l=\{raw, ref\}$. $raw$ represents the raw customer's geological position, whereas $ref$ represents the position of the reference points. The reference point in this study is the location of the target shop.

\subsubsection{Features of aligned location data:}
We also note that shop categories are annotated in our location data. Here, we formally denote our aligned location data, $G\in \mathbb{R}_{+}^{ U \times V \times N \times M}$ and $g_{u, v, c, t} \in T$. $g_{u, v, c, t}$ represents the number of visits to a shop of category $c$, and its aligned latitude and longitude are $(u, v)$, and $t$ represents its experimental group (Treatment or Control). 

\subsubsection{Specification of spatial difference of the shopping trajectory:}\label{method:traject_calc}

For \textbf{RQ1}, we investigate if the customer in the treatment group shows a different spatial trajectory from the control group, modeling a spatial relationship between shop visiting behaviors and the distance from the target shop.

Because spatial analysis can easily experience difficulties such as spatial dependency or auto-correlation among regions, we employ geographically weighted regression (GWR)~\cite{brunsdon1998geographically}, a weighted regression model that incorporates spatial heterogeneity. We specify the following geographically weighted logistic regression model: 

\begin{equation}
y_{i}=\sum_{j=0}^{m} \beta_{bw, j}\left(u_{i}, v_{i}\right) x_{ij}+\varepsilon_{i},
\end{equation}
where $y_{i}$ is a binary dependent variable that represents the dominance of either group; $x_{i,j}$ is a vector of independent variables for area $\left(u_{i}, v_{i}\right)$; $\beta_{bw, j}$ represents that for each feature $j$, we use a different bandwidth $bw$ for weighting. Dependent variables, here is whether the treatment group dominates are $\left(u_{i}, v_{i}\right)$. For the independent variable at position $i$, $x_{i,j}$, we use the distance from the reference points and shop category shares. 

We set $y_{i}$ as 1 when the customers in the treatment group visit a block $\left(u_{i}, v_{i}\right)$ more frequently than the customers in the control group, and otherwise 0. Because the number of observed data points for each group is different, we normalized the data as described in Appendix~\ref{sec:norm}.

\subsection{RQ2: Travel Distance Differences After the Target Shop Visit Day}
\label{method:travel_distance}
When a customer changes their trajectory, it will be of paramount interest whether that alternation is a permanent or tentative. We study the shopping trajectory after the day when the customer visits the target shop. To trace the customer trajectory, we calculate the travel distances nearly 3 days before and after the first visit day, denoting the travel distances of user $i$ at day $s$ as $d_{i,s}$. Here, $d_{i,s}$ represents the position from the day that customer $i$ visits the target shop, and we focus on the three days before and after the target shop visit day (i.e., $s = \{j\}_{j=-3}^{3}$). For example, $d_{i,s=0}$ is the travel distance of the day that user $i$ visited the target shop. We calculate the travel distance after the visit to the shop between the control and treatment groups using the following regressions. 

\begin{equation}
d_{i,s} =\alpha +\operatorname{Aft}_{t} \cdot T_{i} \beta + X_{i, s} {\large \gamma} + \epsilon_{i, s} 
\end{equation}

, where $\alpha$ denotes an intercept; $T_{i}$ the dummy variable that takes 1 when customer $i$ is in the treatment group and otherwise 0; ${Aft}_{s}$ becomes 1 after the shop visit (i.e., $s > 0$); otherwise, 0; $ X_{i s}$ represents the control variables that incorporate customer fixed effects; $\epsilon_{i s}$ denotes the error term. 

In this model specification, $\beta$ represents the average differences in the travel distance between the voluntary (control) and invited (treatment) group after the shop visit day. For control variables, $X_{i s}$, we use several sets of fixed effect variables that might affect travel distance as follows: the day of week, the day, the advertisement, and individual fixed effect. We will attempt several settings of $X$ and discuss how it affects the estimation of the travel distance differences. For instance, having a fixed individual effect implies that we control the customer's difference, such as their travel distance preference. Especially, when we incorporate the individual and time fixed effects, this regression model becomes the same as the framework of difference-in-difference, which is a widely used causal inference technique not only in social science~\cite{angrist2008mostly, lechner2011estimation, card1993minimum} but in a recent computer science study~\cite{liPodcast2020, kusmierczyk2018causal}.

\subsection{RQ3: Revisit Effect of Advertisements}~\label{method:re-visit}

We are also interested in studying the long-term effect of the advertisements, the revisiting effect. In other words, we intend to know the ratio of the customers that the advertisement makes them revisit. To study this effect, we subtract the revisit ratio in the control group from the treatment groups' revisit ratio. Because the control group did not receive an advertisement, the subtracted revisit ratio can be interpreted as the effect of the advertisement. In addition to the simple calculation, we conduct meta-analysis to consider the advertisement differences.

\subsubsection{Calculating the revisiting by the advertisement: } As in Sec.~\ref{sec:study_design}, We assume that there are two types of customers: Induced and organic customers. Induced customers are the customers induced by the advertisements. Meanwhile, there are customers who visit the target shops regardless of the advertisements, voluntary customers.

The revisit probability $Pr(RV)$ can be formalized as follows: 
$
Pr(RV) =  Pr(RV_{organic}) + Pr(RV_{induced})
$
where $Pr(RV_{organic})$ and $ Pr(RV_{induced})$ denote the revisit ratios of the organic and induced customers, respectively~\footnote{
Note that
$\text{\#revisit customers}= \text{\#organic revisit customers} $+$ \text{\#induced revisit customers}$, and therefore
\newline
$\text{Pr(RV)}=\frac{\text{\#revisit customers}}{\text{\#first visit customers}} = \frac{\text{\#organic revisit customers}}{\text{\#first visit customers}} +\frac{\text{\#induced revisit customers}}{\text{\#first visit customers}}$
}.
With this formalization, we obtain $Pr(RV_{induced})$. As a definition, all of the re-visiting customers in the control group do not receive an advertisement. Therefore, we estimate $Pr(RV_{induced})$ as~\footnote{Note that this is not the average treatment effect (ATE) because the re-visit is only defined for the population who visit the shops at least one time.} 

\begin{eqnarray}\label{eq:revisit_emp}
Pr(RV_{induced}) &=& Pr(RV) - Pr(RV_{organic})\\
&=& Pr_{treatment}(RV) - Pr_{control}(RV)\\
&=& E[Revisit|T=1] - E[Revisit|T=0]
\end{eqnarray}

\subsubsection{Controlling the differences among the advertisements: }
Although estimating Equation~\ref{eq:revisit_emp} is straightforward, it might contain some biases because our data set consists of the experiments for the various advertising campaigns. Differences among them can yield heterogeneous effects on revisiting, resulting in an imprecise estimation. To overcome this potential problem, we employ the model of the variance of the experiments and study the overall effect of the experiments. To this end, we conducted a meta-analysis using the Mantel-Haenszel method with random effects, which is a popular method in clinical research~\cite{Mathias2019}. This meta-analysis method returns a weighted risk ratio assuming the variance among the experiments as random effects. To make our results comparable, we will report the odds ratio of revisiting for the invited customer (treatment) group with the voluntary (control) group by the following two methods: the direct calculation described in Equation~\ref{eq:revisit_emp} and the meta-analysis.

\subsection{Causal Relationship Prediction}
\label{method:uplift}
Finally, we propose a prediction model for high-value customers. In our analysis, we will find the customers who revisit the target shops after the first visit. These customers are supposed to bring more profit to the business than the customers who visit only once. Therefore, a model to predict the revisiting effect is an essential issue in marketing. 

The most natural solution for this prediction task might be to find the factors associated with customers' revisit behaviors. For example, logistic regression analysis will promptly find the variables that correlate with the revisiting behavior. Several marketing studies conventionally perform this type of post-experiment analysis. However, the challenge that advertisers face in the real situation is the efficient distribution of advertisements to potential customers under the budget constraint. From this point of view, we should predict the degree to which an advertisement improves the probability of revisiting. The prediction model for this aim, in particular, is a model to predict the difference between the revisit probability when we show a customer $i$ the advertisement and when we do not. This difference is known as {\it lift effect} in the causal machine learning literature. 

\subsubsection{Definition and estimation of lift effect: }
To predict the lift up effect, we employ the {\it uplifting model}, the state-of-art counterfactual machine learning models~\cite{gutierrez2017causal,rzepakowski2012decision,zaniewicz2013support,kawanaka2019}. The ultimate aim of the uplifting model is to predict customer $i$'s lift effect of the treatment $\tau_{i}$, which is defined as follows: 
\begin{equation}\label{eq:uplift_def}
\tau_{i} = \operatorname{P}\left(Revisit_{i}=1 \mid T_{i}=1\right)-\operatorname{P}\left(Revisit_{i}=1 \mid T_{i}=0\right).
\end{equation}

The first term on the right-hand side of Equation~\ref{eq:uplift_def} is the probability that customer $i$ revisits the target shop if it receives the advertisements, whereas the second term is the probability that customer $i$ revisits the target shop if it does not receive the advertisements. Intuitively, Equation~\ref{eq:uplift_def} calculates the degree that the advertisement \textbf{lifts} the probability of customer $i$ revisiting.

Apparently, we could observe only one of the two. Therefore, we estimate a function that predicts the $\tau$ value for a given customer, utilizing the experimental data. To predict $\tau$, we want to have a function as follows: 
\begin{equation}\label{eq:uplift_two}
\tau\left(X_{i}\right)=\operatorname{P}\left(Revisit=1 \mid X_{i}, G=T\right)-\operatorname{P}\left(Revisit=1 \mid X_{i}, G=C\right)
\end{equation}
where $X_i$ denotes the feature vector of customer $i$ discussed in~\ref{sec:upliftmodel_feature}, $G={T, C}$ represents whether a customer receives a treatment (i.e., receives an advertisement) $T$ or not $C$. The first term describes the probability of customer $i$'s revisiting given $X_i$ when they are in {\it treatment groups} and the second term is the probability of revisiting even when they do not receive an advertisement. 

We could estimate the first and second term separately from the treatment and control group data, but it requires training two different models for a single prediction function. To avoid this cumbersome task, we use the reverted model proposed by~\cite{zaniewicz2013support}, which is a transformed version of Equation~\ref{eq:uplift_two}; that is, introducing the indicator $Z$ as follows:

\begin{equation}
    Z=\left\{
    \begin{array}{ll}
        1 & \text{ if G=T and Revisit=1} \\
        1 & \text{ if G=C and Revisit=0} \\ 
        0 & \text{ otherwise. }
    \end{array}
      \right.
\end{equation}

As~\cite{zaniewicz2013support} explains, $Z$ takes 1 if the assignment $G$ is ideal. In other words, we want to see the customers with the treatment revisit, and the customers without the treatment not revisiting. Note that $Z$ takes 0 when a treated customer does not revisit, or a non-treated customer revisits. We will not gain any treatment for both cases. Under the assumption that the treatment assignment is random, we obtain \footnote{The detail of the calculation can be found at~\cite{zaniewicz2013support}}

{\footnotesize
\begin{eqnarray}
\tau\left(X_{i}\right)&=&
\operatorname{P}\left(Revisit=1 \mid X_{i}, G=T\right)-\small{\operatorname{P}\left(Revisit=1 \mid X_{i}, G=C\right)}\\
&=&2 P\left(Z_{i}=1 \mid X_{i}\right)-1.
\end{eqnarray}
}
Therefore, we will estimate the function $\tau(X)$ through \scalebox{0.9}{$P\left(Z=1 \mid X\right)$}. 

\subsubsection{Evaluation of the uplift model: }
To evaluate the uplift model, we will study the mechanism through which the advertisement distribution based on the model increases the uplift effect compared to the random advertisement distribution. Following practical conventions~\cite{rzepakowski2012decision,radcliffe2012real}, we use the area under the uplift curve (AUUC) for the evaluation. The AUUC is defined as follows:
\begin{equation}
AUUC =\int_{k = 0}^1 f_{1}(k)-f_{0}(k) dk,
\end{equation}
where $f_{o}(k),\ o \in \{1, 0\}$ denotes the lift effect of the $k$th sample in the ordered sample $o$. For $o=1$, we sort the sample based on the uplift effect prediction (descending). In contrast, we arrange the sample in a random manner when $o=0$. Intuitively, the AUUC describes the uplift gain when we distribute the advertisements from the highest uplift users to the lowest uplift users.

\section{results}

This section will report our results that answer the three research questions proposed in the introduction section. First, we focus on the customer trajectory near the target shop during their first target shop visit (\textbf{RQ1}). In addition to presenting the actual customers' trajectory, we model the customers' spatial behavior and report the model prediction to show the distinctions between the two groups. Furthermore, we trace the trajectory three days after the first target shop visit day and calculate the travel distances to observe the persistence of the advertisement effects (\textbf{RQ2}). Then, we will study the revisiting effect of advertisements (\textbf{RQ3}), comparing the revisiting ratio between the control and treatment groups. Finally, with the finding in \textbf{RQ3}, we propose a counterfactual machine learning model for marketing strategy based on our analysis. We use the uplift model to predict the revisit effect and a strategy for delivering the advertisement to maximize the revisit effects.

\subsection{RQ1: Shopping Trajectory Differences}
\label{res:rq1}
Our first research question asks; "Do the customers invited by the advertisement show different shopping spatial trajectories on the target shop visit day?" To investigate this trajectory difference, we compare the trajectories of the voluntary visit customers (the control group) and the invited visit customers (the treatment group) during their first target visit day.

\subsubsection{Customer location history near the target shop}
To answer \textbf{RQ1}, we study the customer location data near the target shop to compare the differences in visit points between the two groups. We first align the customer location history to map their trajectories into the same coordination as in Sec.~\ref{method:coordination}. After coordination, we focus on an area of 2km from the target shops in radius, and calculate the number of visits in each grid (111$m^{2}$), as discussed in Sec.~\ref{method:traject_calc}. Figure~\ref{fig:tre_con_shop} shows the area near the target shop with a label that represents the group that dominates a particular  area. Not compelling but Figure~\ref{fig:tre_con_shop} alludes to the fact that the customers in the control group are dense near the shop and the treatment group are dominant in the outward area.

\subsubsection{Customer spatial trajectory modeling: }
To conclude the hypothesis proposed above, we will employ a GWR regression to model a spatial relationship between the distance from the target shop and the dominant area. We first model a simple regression where we only use the distance from the shop. The results of the regression in Table~\ref{table:gwr} demonstrate a positive relationship between distance and treatment group dormancy. The further from the target shop a place is located, the more likely the treatment group dominates that place. These results are qualitatively the same even when we add some control variables that might affect the area preference. In the second model, we add the ratio of the shops in each category, shopping and food, and found a positive relationship between distance and treatment group dominance.

\begin{figure*}
\captionsetup{justification=centering}
\begin{multicols}{3}
    \includegraphics[width=\linewidth, height=0.8\linewidth]{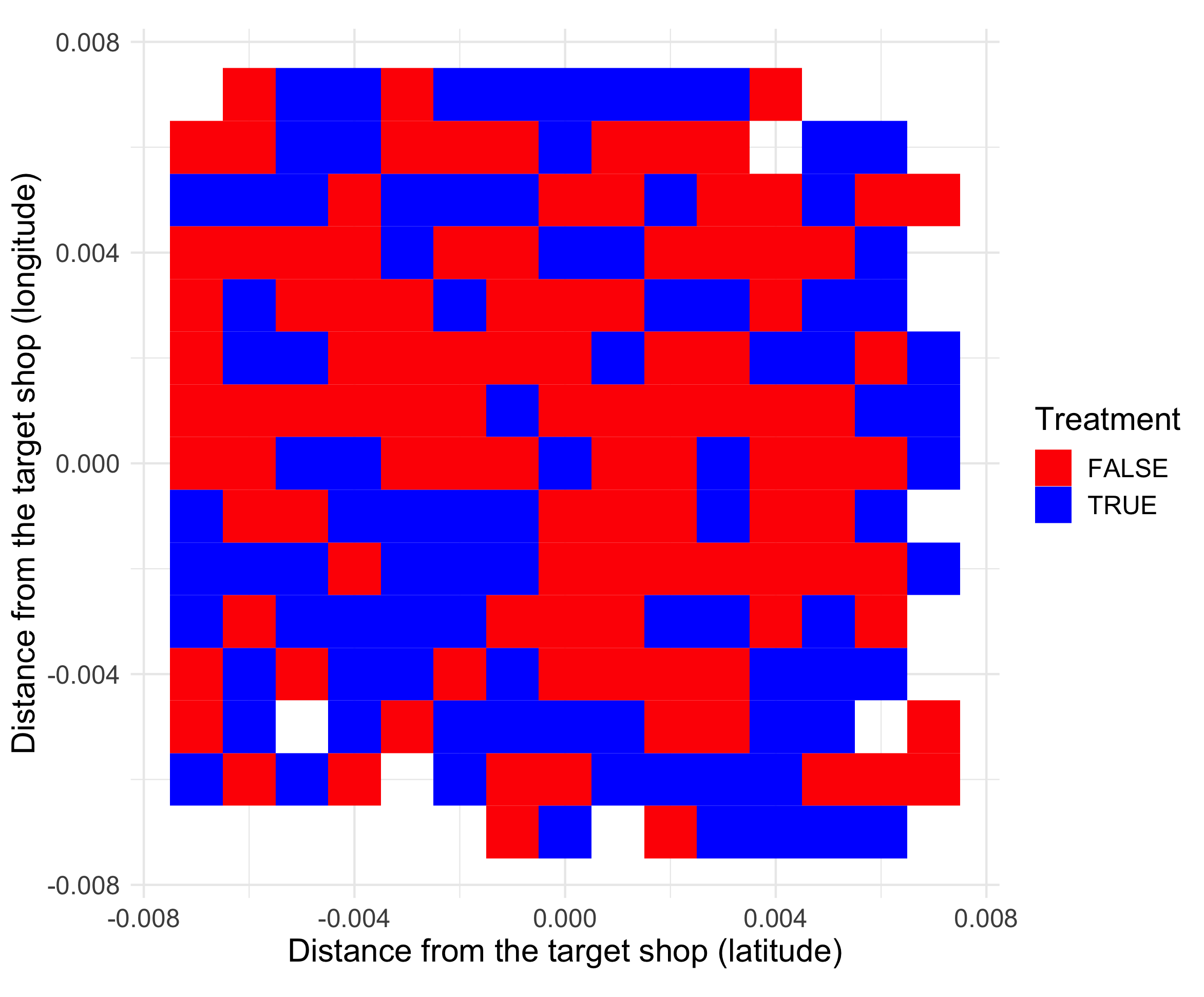}
 \caption{Dominant areas around the target shop}
\label{fig:tre_con_shop}\par 
    \includegraphics[width=\linewidth, height=0.8\linewidth]{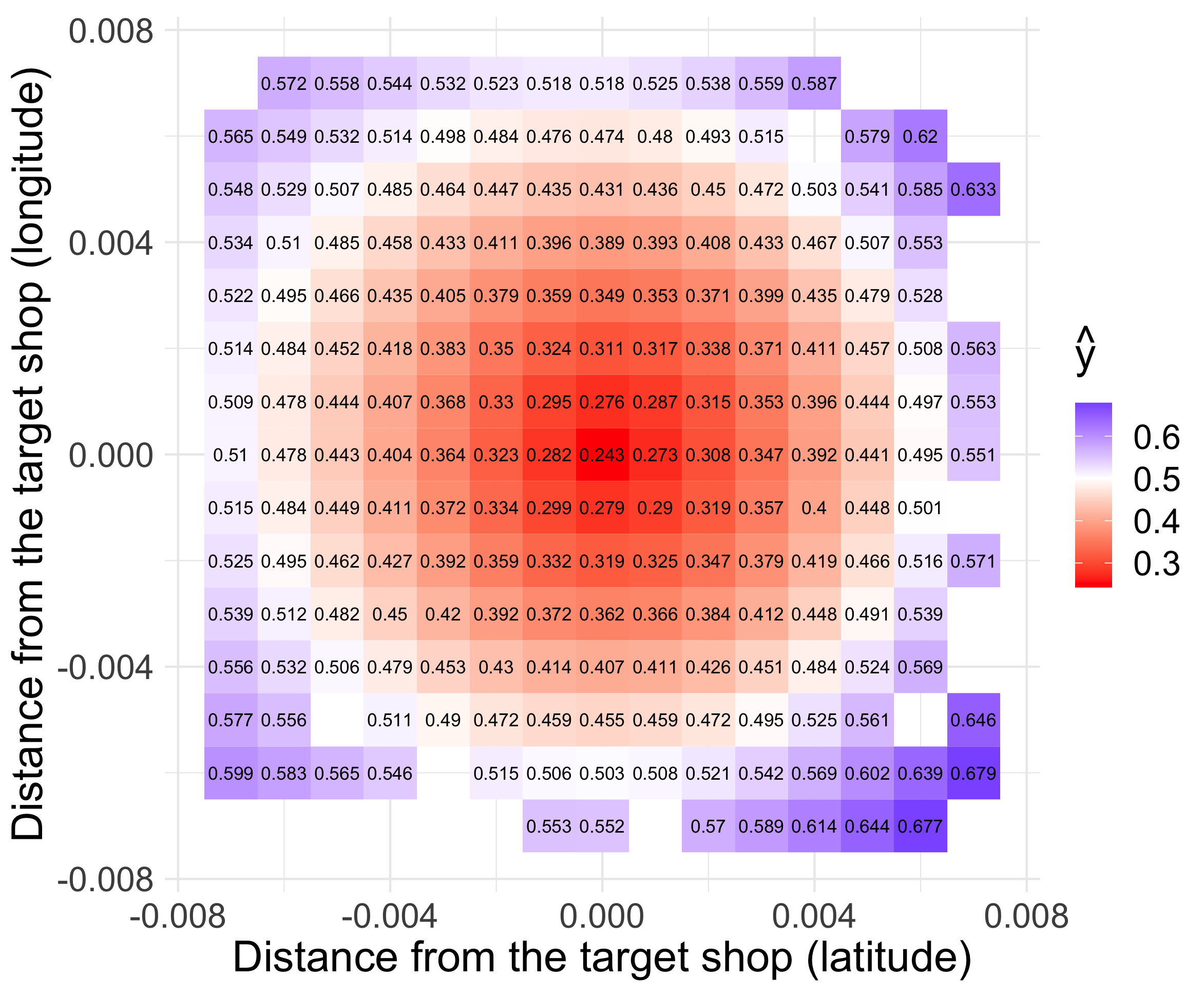}
 \caption{GWR prediction \\(simple model, Model 1)}
\label{fig:gwr_prediction_sim}\par 
    \includegraphics[width=\linewidth, height=0.8\linewidth]{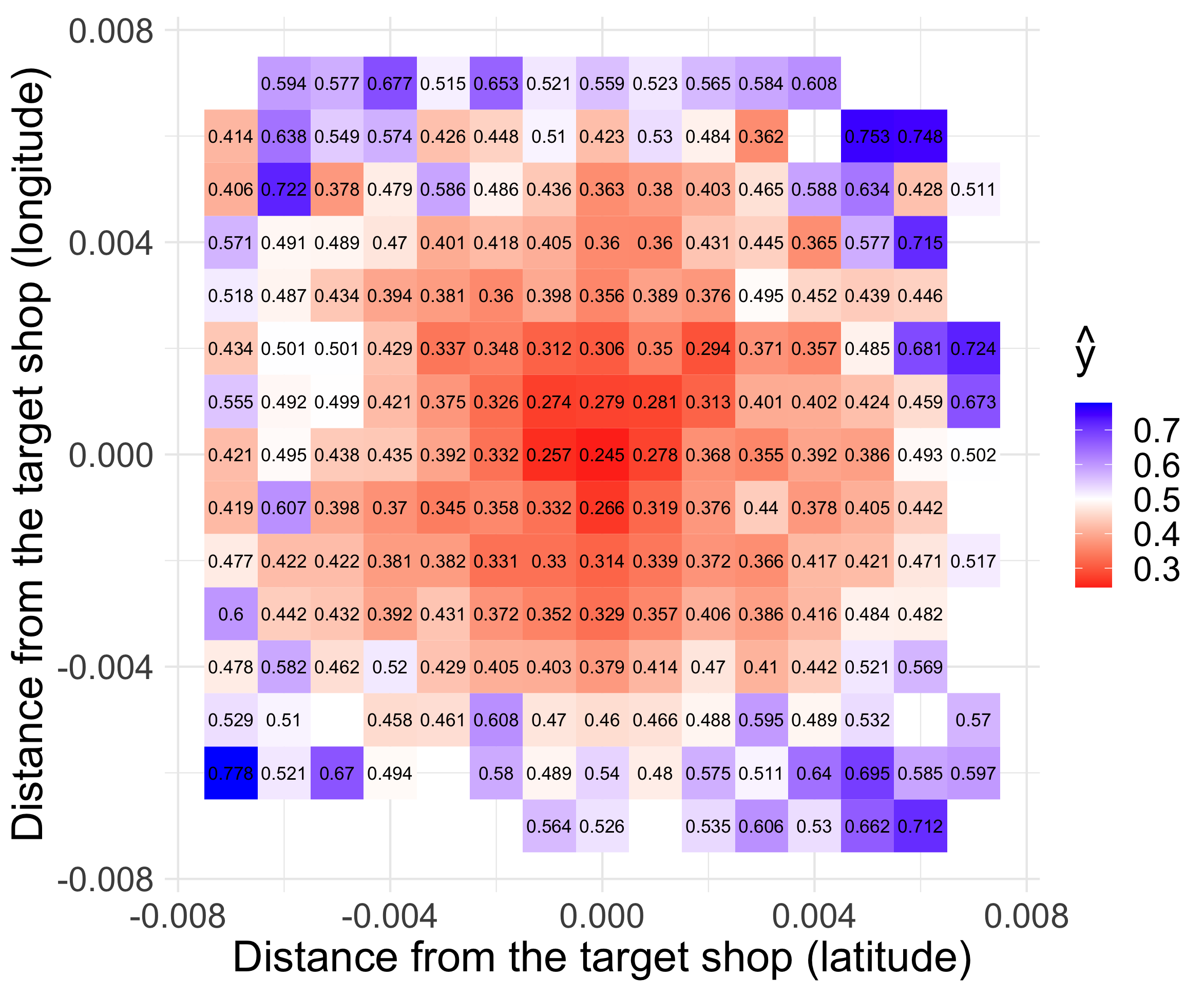}
\caption{GWR prediction \\(multivariate model, Model 2)}
\label{fig:gwr_prediction_mul}\par 
\end{multicols}
\begin{minipage}{\columnwidth} %
\vskip 0.03in
{\footnotesize {\it Note: }We align the target shop to the center of the plot and then split the area around the target shops by grids (the area of a square is about $111m \times 111 m$). \textbf{Figure~\ref{fig:tre_con_shop}} shows the areas the invited customers (blue,treatment) group dominant and the voluntary customers (red, control) group dominant. We calculate which blocks are dominant following the procedure described in Sec.~\ref{method:traject_calc}. \textbf{Figure~\ref{fig:gwr_prediction_sim}} describes the prediction by the estimated GWR model (Model 1 in Table~\ref{table:gwr}). The figure shows that the model predicts that the invited customers dominants the outer areas (Blue), whereas the voluntary visit customer condence around the target shop (Red). \textbf{Figure~\ref{fig:gwr_prediction_mul}} provides the prediction by the GWR model with multiple independent variables (Model 2 in Table~\ref{table:gwr}). The second model also predicts the dominance of the outer areas by the treatment groups.\par}
\end{minipage}
\end{figure*}

This finding becomes more evident as we observe the prediction results using the estimated models. Figure~\ref{fig:gwr_prediction_sim} depicts the prediction result where the control group dominates the center areas, whereas the treatment group dominates the outer areas. We also found a similar result when we used the second multivariate model. Note that the aim of these prediction models is not to prediction the behavior, but to abstract the important factors from the customers' trajectories.

These prediction results contrast the trajectories of customers with two different motivations for visiting the target shop. Voluntary visitors gather around the area near the target shops. Meanwhile, the invited customers move to wider areas. The customers who shop in a wide area would bring more benefits to the shops near the target shop. In other words, the invited customers bring a larger externality in the sense that the shops near the target shop benefit from them, even though the campaign did not advertise these nearby shops.

\begin{table}[t]\small
\caption{Results: GWR model}
\label{table:gwr} 
\begin{center}
    \begin{tabular}{lccl}
            & \textbf{Parameter} & \textbf{Std. Err.} & \textbf{P-value} \\
    \toprule
    \multicolumn{4}{l}{\textbf{Model 1: Simple regression}}\\
    Dist from the shop &  170.262  & 72.098   & 0.0182** \\
    Intercept          &  -1.082   &  0.423   & 0.0105** \\
    \midrule
    \multicolumn{4}{l}{\textbf{Model 2: Multivariate regression }}\\
    Dist from the shop & 159.4583  &  72.7248 & 0.0283** \\
    Food shop pct      & -1.2004   &  0.6966  & 0.0848*  \\
    Shopping pct       & -0.9626   &  0.7061  & 0.1728   \\
    Intercept          & -0.1784   &  0.6670  & 0.7892   \\
    \toprule
    \end{tabular}
\vskip 0.03in
\begin{minipage}{\columnwidth}%
{\footnotesize {\it Note: } The estimation results of the logistic GWR. We estimate the model that predict whether a given block is dominated by the invited visit customers or not. In both two models, we find that the distance from the target shop has positive coefficient. In other words, the invited customers tends to dominate the outer areas than the voluntary visit customers, implying that the invited customers travel wider area. While \textbf{Model 1} is the simple regression model, \textbf{Model 2} incorporates the shop category ratio of that area as the control variable that might affect the travel decision. The number of stars * represent statistical significance: * p-value$< 0.1$; **$< 0.05$; ***$< 0.001$.
\par}
\end{minipage}
\end{center}
\end{table}

\subsection{RQ2: Persistence of Trajectory Difference}

After \textbf{RQ1}, one interesting question would be; ``How long do these trajectory differences persist? (\textbf{RQ2})''. Such an analysis will enable us to know the duration of the behavior difference between the voluntary and invited customers found in \textbf{RQ1}.

To understand the persistence of the advertisement externality, we will use the daily travel distances of the customers before and after the first target shop visit day. Because we found that the invited  (treatment) group travels wider areas than the control group, the travel distances of each group should be a suitable indicator of such externality. In this analysis, we calculate the daily average travel distance for each group. There are several confounding factors that might affect their travel distance such as advertisement differences, individual differences, and differences among days. We control these differences by incorporating control variables, as discussed in Sec.~\ref{method:travel_distance}.

\subsubsection{First look: }
We first calculate the travel distance difference among the two groups for each day, controlling the advertisement differences by the advertisement fixed effect. We run the regression analysis and report the coefficients of the day-dummy in Figure~\ref{fig:dist_visit_trans}. The figure shows that the travel distances increase on the target shop visit day, and the distance differences remain relatively higher than the day before the visit. Even though the confidence intervals of Days 1 and 2 are on the zero bound, we are not sure if these differences are robust. For this first analysis, we only control for differences among the advertisements, but each individual might respond differently. We therefore need to conduct a regression analysis modeling these differences to determine if this finding is robust.

\subsubsection{Analysis with elaborated models: }
We attempt four settings of the control variables and report the results in Table~\ref{table:did}. The table calculates the day average travel distances three days after the target shop visit day~\footnote{The target shop visit day is not included.}. Consequently, we find that the invited (treatment) group travels a longer distance even after the shop visit day. We estimate Model 1 as a baseline that only incorporates the dummy variables of the advertisements. For Model 2, we also incorporate the day-of-week dummy into the models. These first two models return similar results in the travel distance differences, suggesting that the day-of-week effect is negligible. 

On the contrary, the last two models with the heterogeneity among the customers (Models 3 and 4) report smaller travel differences than the first two (Models 1 and 2), indicating that there is a certain heterogeneity among the customers and its overestimates of travel distances. While detecting the overestimation in the first look analysis, our analysis of the persistence of trajectory difference demonstrates consistent results that the treatment group travels longer distances even after the target shop visit day.

\begin{table*}[ht]\small
\begin{center}
\caption{Travel Distance Differences: the voluntary visit customer VS the invited customer}
\label{table:did}

      \begin{tabular}{lccc cccc}
      
  & { \textbf{Parameter}}
  & { \textbf{Std. Err.}}
  & { \textbf{P-value}}
  &
  & {\textbf{Parameter}}
  & { \textbf{Std. Err.}}
  & { \textbf{P-value}}\\ 
\toprule

& \textbf{Model 1: } & & &
& \textbf{Model 3: } & &\\
\cline{2-4} \cline{6-8}
\textbf{Difference} &    2.5185  &    1.2085 &  0.0372** & & 2.3672 & 1.2068 &  0.0498** \\
\textbf{Baseline}   &    37.106  &    3.0694 &  <0.001*** & & 35.216 & 0.2529 &  <0.001***  \\
\cdashline{2-4} \cdashline{6-8}
&Ad\checkmark& Customer &  DoW 
&&Ad \checkmark& Customer\checkmark &  DoW\\
&\multicolumn{1}{c}{Day}&                  &&              &\multicolumn{1}{c}{Day}&       &\\

& \textbf{Model 2: } & & &
& \textbf{Model 4: } & &\\
\cline{2-4} \cline{6-8}
\textbf{Difference}&    2.4843 &  1.2021 &  0.0388** & & 2.3465 &    1.3174 &  0.0749*  \\
\textbf{Baseline}   &    35.631 &  3.1176 &  <0.001*** & & 33.779 &    0.2843 &  <0.001*** \\
\cdashline{2-4} \cdashline{6-8}
&Ad\checkmark & Customer &  DoW \checkmark 
&&\multicolumn{1}{c}{Ad \checkmark}& Customer\checkmark&  DoW\checkmark\\
&\multicolumn{1}{c}{Day}&                  &&              &\multicolumn{1}{c}{Day\checkmark} &       &\\
\midrule
\end{tabular}

\begin{minipage}{\columnwidth} %
{\footnotesize{\it Note: } 
Travel Distance Differences estimated by the regression models described in Sec.~\ref{method:travel_distance}. All of the results shows that the invited customers travel longer distance than the voluntary visit customers about 6.8\% ($\approx 2.42/35.42$). We calculate the difference between e voluntary visit customer and the invited customer after the shop visit. \textbf{Difference} is the estimated average difference using the voluntary visit customers' average travel distance as a baseline (i.e., the invited customer - the voluntary visit customer). \textbf{Baseline} is the average travel distance of the voluntary visit customer. In each model, we try the different set of fixed effects to control the confounding factors that may affect the travel distance. \textbf{The check mark $\checkmark$} below the estimation results indicates that the model incorporates which fixed effect; \textbf{Ad}: Advertisement fixed effect; \textbf{Customer}: Customer-level fixed effect; \textbf{Dow}: Day-of-week fixed effect; \textbf{Day}: day-level fixed effect (the day position from the target shop visit day). The number of stars * represent statistical significance: * p-value$< 0.1$; **$< 0.05$; ***$< 0.001$.
\par}
\end{minipage}
    \end{center}
\end{table*}

\subsection{RQ3: Revisit Odds Differences}

Finally, we expand the time horizon of the analysis and ask \textbf{RQ3}, studying if the customers revisit the shop within 4 months after the visit. To answer this question, we calculated the re-visitation ratio difference between the treatment and control groups, studying if a user revisits the shop within 4 months after the visit. We also present a causal machine learning model to predict the re-visitation effect of the advertisement and discuss the marketing strategy for re-visitation.

\subsubsection{Revisiting ratio difference: }
As discussed in Sec.~\ref{method:re-visit}, we interpret the differences among the groups in the revisit ratio as the revisits invited by the advertisements. We calculate the odds ratio of revisiting for the invited customer (treatment) group with the voluntary (control) group. Because our data consists of multiple advertisement campaigns, the effect of advertising may vary. We therefore also present the odd ratio using Mentel-Haenszel methods to take into account the advertisement effects' heterogeneity.

\begin{figure*}\small
\begin{multicols}{3}
    \includegraphics[width=1.1\linewidth, height=0.48\linewidth]{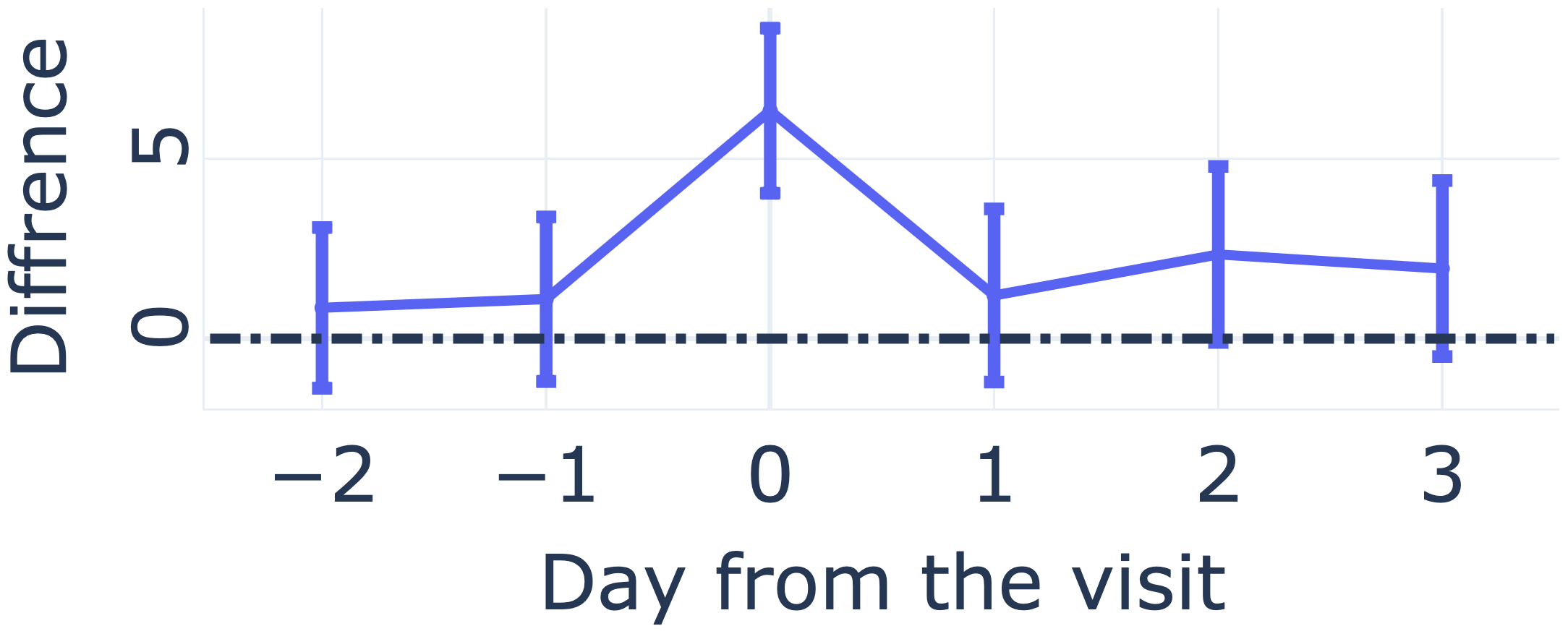}
\caption{Travel distances difference (km) before \& after the shop visit day}
\label{fig:dist_visit_trans}\par 

    \includegraphics[width=0.98\linewidth, height=0.55\linewidth]{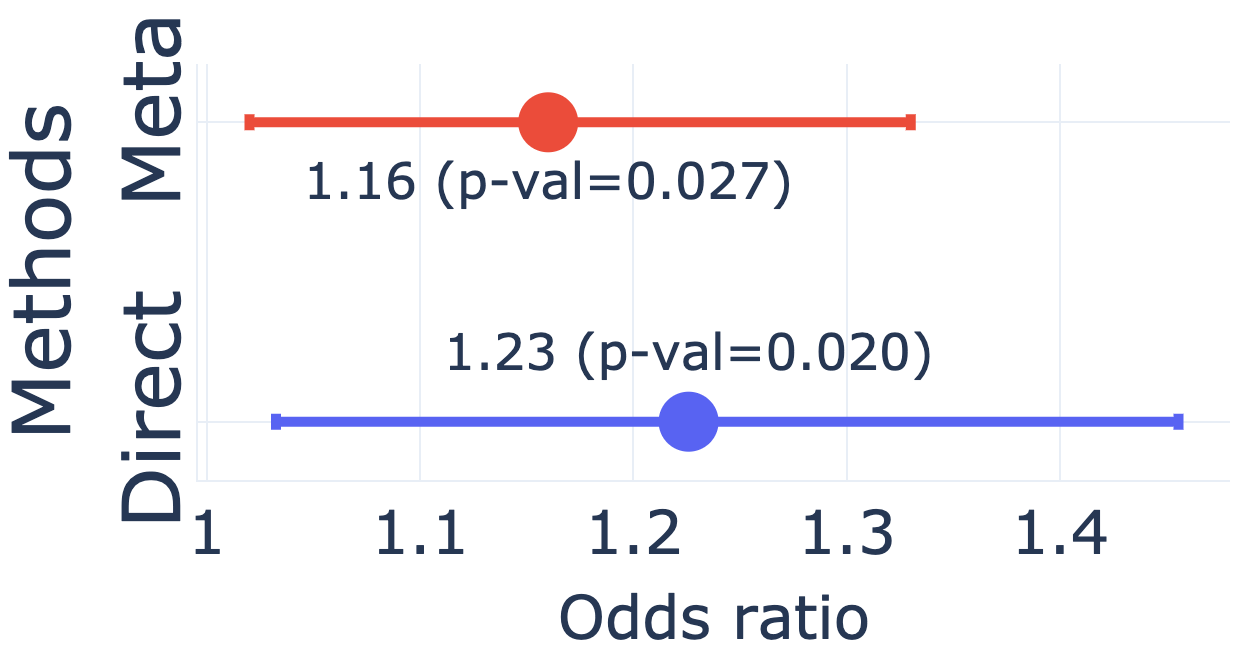}
\caption{Odds ratio of revisiting}
\label{fig:revisit_odds}\par 

    \includegraphics[width=1.1\linewidth, height=0.6\linewidth]{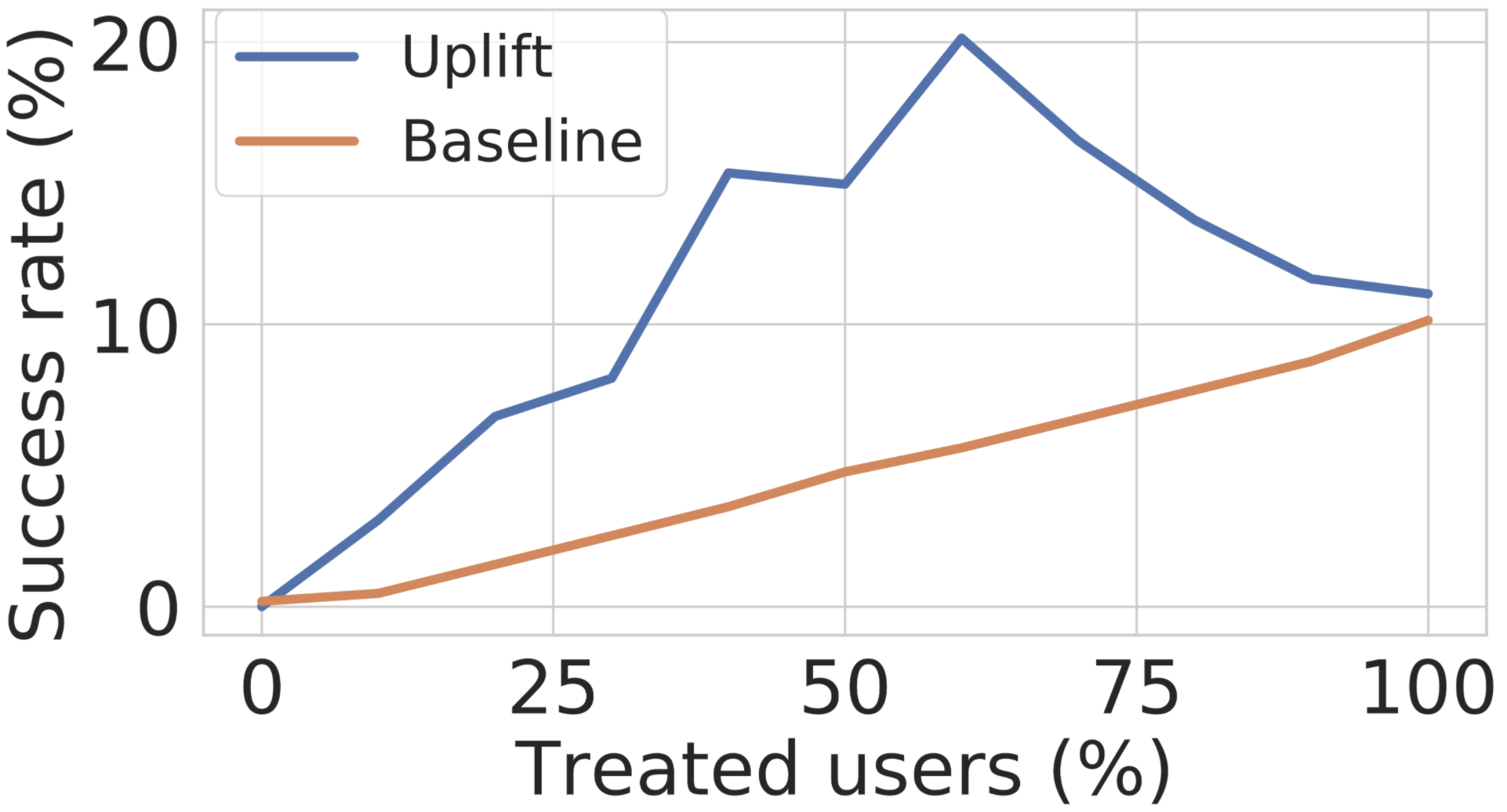}
\caption{Uplift curve}
\label{fig:uplift_crv}\par 

\end{multicols}
\begin{minipage}{\columnwidth} %
{\footnotesize {\it Note}: \textbf{Figure~\ref{fig:dist_visit_trans}} describes the transition of the travel distance differences (km) between the voluntary and invited customers (the error bars represent 95\% confidence interval.). \textbf{Figure~\ref{fig:revisit_odds}} compares the odds ratio of revisiting for the invited customer (treatment) group with the voluntary (control) group by the two difference methods: Direct and Meta analysis. \textbf{Direct}: The odds ratio of the whole sample. \textbf{Meta}: The odds ratio by the Mantel-Haenszel method that controls the differences between the advertisements. \textbf{Figure~\ref{fig:uplift_crv}} shows the marketing strategy to maximize the lift effect of revisiting by the advertisements. The figure shows that the revisiting is maximized when 60\% of the customers receive the advertisements, which means that.}
\end{minipage}
\end{figure*}

Figure~\ref{fig:revisit_odds} reports the results of the two different odds with 95\% confidence intervals. In both calculations, the odd ratios are greater than 1, implying that the customers who received an advertisement are more likely to revisit than those who visit a shop voluntary. In other words, the advertisements encourage revisiting of the customers who do not revisit without the advertisements. Figure~\ref{fig:revisit_odds} also shows that the odds estimation by the meta analysis has a narrow confidence interval than the estimation using the direct method. This overestimation by the direct methods suggests that the revisit effect varies among the advertisements. However, after modifying these differences, the odd ratio is still greater than 1 and statistically significant, implying that the advertisements increase the revisit ratio.

\subsection{Predicting the Revisiting Lift Effects} 
We found that the advertisement encourages the customers to revisit the target shop. A natural next goal is to study the way to distribute the advertisement to maximize the revisiting effect. As we discussed in Sec.~\ref{method:uplift}, this goal is not achievable only without modeling the effects that the advertisement lifts (increases) the probability of revisiting a given customer. To this end, we use uplift modeling to predict the revisiting effect for each customer, and discuss a reasonable advertisement strategy. We split the data set into three and conducted feature selection optimization, train model, and predict the uplift with each subset of the data. With the trained model, we study what features are associated with the revisiting effect prediction.

To construct the uplift model, we first determine the best combination of features to maximize the uplifting effects (i.e., the AUUC) from the set of features described in Sec.~\ref{sec:upliftmodel_feature}. A number of candidate features for the model include customer profiles and location information. The best set of features is useful not only for improving the prediction accuracy but also in the sense that it enables us to know the features that are associated with the revisiting behavior. We conduct feature selection optimization via the state-of-the-art framework for hyper-parameter optimization, Optuna~\cite{akiba2019optuna}. The feature optimization selects 60 features that consist of visit place histories, demographic information such as age, and distance from home to the target shops. 

\begin{figure}
    \centering
    \includegraphics[width=0.4\columnwidth, height=0.4\linewidth]{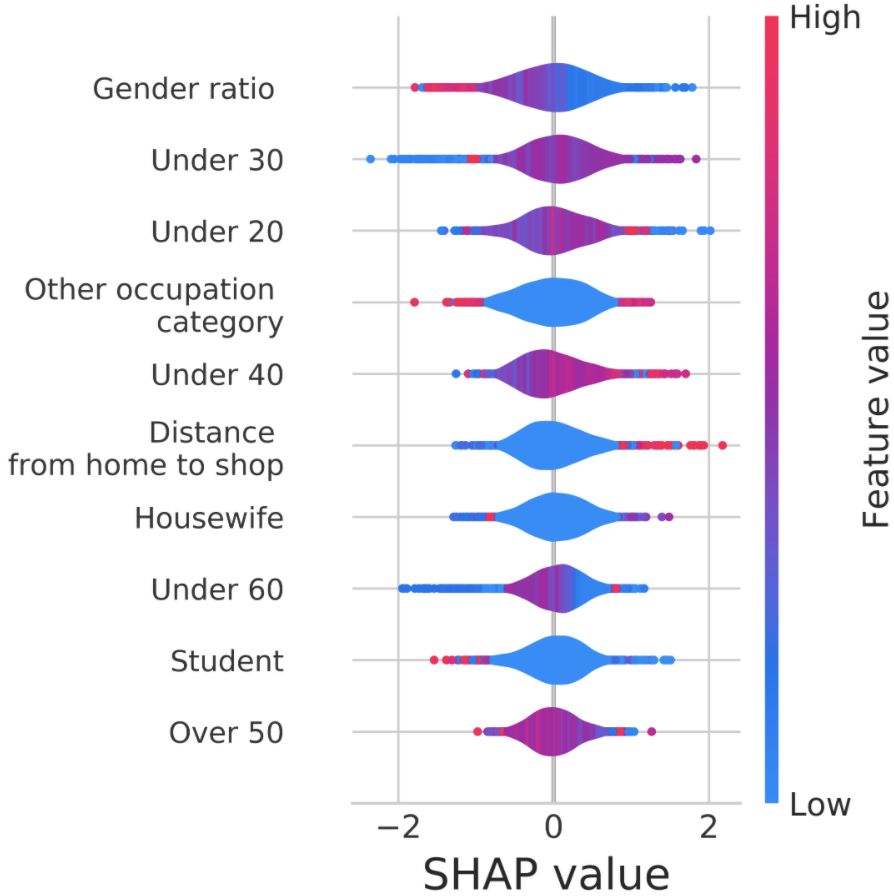}
\caption{The distribution of the SHAP values of the top 10 higest mean SHAP value features}\label{fig:sharp}
\par 
\begin{minipage}{\columnwidth} %
\vskip 0.03in
{\footnotesize {\it Note:}
 The top 10 features by the SHAP value by~\cite{shap2017} for the uplifting model. The descriptions of the features are in Appendix~\ref{app:shap_features} Except for {\it Distance from home to shop}, the other demographic variables represent the probability that a customer in that demographic category. For example, the high value of \textbf{Other occupation category} means the prediction model returns the high probability that the customer is not in the specific occupation category. The color represents the value of the feature (Reg: high, Blue: low). A high SHAP value represents a positive impact on the lift effect, and vice versa. For instance, the figure shows that the customer who lives the spot with fewer woman residents contributes to the high lift effect predictions. Overall, this SHAP value analysis suggests that younger male persons tend to have a higher lift effect.
\par}
\end{minipage}
\end{figure}

With the features selected by the Optuna~\cite{akiba2019optuna}, we use the XGboost classifier~\cite{chen2016xgboost} to estimate our uplift model and train it on the train dataset \footnote{max\_depth: 10, n\_estimators:100}. To understand each feature's contribution to the prediction of the lift effect, we utilize the SHapley Additive exPlanations (SHAP) value~\cite{shap2017}. The SHAP value represents the degree of the contribution of each feature on the prediction~\footnote{Online version of \cite{molnar2019interpretable} provides a good introduction of SHAP value}. We chose the top 10 features based on the mean absolute SHAP values~\footnote{We report the decriptions of the features and the absolute mean SHAP values of them in Appendix~\ref{app:shap_features}.}. A high SHAP value of a feature implies that a high value of that feature has a positive effect on a lift effect prediction and vice versa. Figure~\ref{fig:sharp} reports the distribution of the SHAP values for the top 10 features, and it suggests the importance of predicted demographic information such as age or occupation. The distribution of the SHAP values of the most three important features suggest that our uplift model predicts a high lift effect for young males. In addition, we find that location information is also important, as represented by {\it Dist from the shop to home}. The Figure demonstrates that the longer distance predicts a high lift effect. This result implies that such customers wouldn't have revisit the shop because they away from the shop, and the O2O advertisements can cause the revisit.

With the trained uplift model, we study the optimal advertisement strategy to maximize the revisit effects. We calculate the cumulative success rate and depict it in Figure~\ref{fig:uplift_crv}. The figure demonstrates the cumulative success rate when we distribute the advertisement to the customers based on the model's predicted uplift score. The y-axis represents the cumulative success rate (i.e., cumulative revisit rate), and the x-axis represents the percentage of the customers who received the advertisement. The cumulative success rate achieves the peak when approximately 60\% of people received the advertisement. Hence, the uplift model's optimal strategy by this model is to distribute to the customers in the top 60\% uplift scores.

\section{Related Work}
This section first reviews prior studies on the human offline trajectory mining in general and then focuses on the offline customer trajectory. Finally, we discuss the application of the mining human trajectory location-based advertisements, including O2O, and we also introduce the important marketing concept in these topics, commercial area.

\subsection{Mining Human Offline Trajectory}
The mining of human offline trajectory has been an essential computer science topic, which focuses on analyzing the user's physical behavior theoretically or empirically to extract useful information from this high-dimensional data. One of the most fundamental interests in this topic is to find frequent trajectory patterns~\cite{liu2011mining, gaffney1999trajectory}, as frequent patterns allow us to categorize users. For example, the frequent trajectory patterns are often used for clustering the users. Accordingly, the researchers study the trajectories' semantics to generate interpretable results~\cite{nanni2006time,lee2007trajectory,ghosh2016thump,zhou2019context}. In recent years, embedding techniques such as word2vec have been used to approximate high-dimensional offline trajectories to lower dimensions~\cite{crivellari2019motion,crivellari2020trace2trace}.

There are several applications of mining offline human trajectory. The first important application is to predict the next offline trajectory of users~\cite{monreale2009wherenext} because of its usefulness not only for the business but also for the public policy for human mobility~\cite{gambs2012next,gao2019predicting}. However, the limits of predictions based on offline trajectories have been discussed in the context of social networks~\cite{song2010limits}. In social network analysis, this prediction problem is defined as edge connection prediction~\cite{althoff2017online,sapiezynski2018offline}, which reveals the importance of the social network. In addition, the users' offline trajectory can predict their demographic information ~\cite{raviPredictingConsumerLevel2018c,zhong2015you,wang2017inferring,hinds2018demographic,bao2016geo}.

\subsection{Customer Offline Trajectory}

Marketing scientists are especially interested in a more specific human offline trajectory and customer shopping trajectory. This is because it conveys the most critical aspect of consumer behavior during their shopping, for instance, the selection of the times of purchase by the customer. From this perspective, marketing scientists have proposed models that incorporate the customer shopping trajectory, such as traditional marketing strategy models or statistical models.

We would like to emphasize that marketing researchers have mainly focused on customers’ in-store offline trajectories in the context of spatial analysis~\cite{bronnenberg2005spatial,j2005spatial}. For instance, some early work on customers' in-shop trajectory study the customers' shopping path in a super market~\cite{larson2005exploratory} 

To understand the in-store trajectory pattern, Hui et al. have deeply exploited customer behaviors. They first modeled the customer shopping path~\cite{hui2009path,hui2009testing}. To investigate the advertisement and shopping trajectory, they conducted field experiments on in-store purchases with coupon~\cite{hui2013the} detailed analysis with video data~\cite{hui2013deconstructing}. A more extended study by ~\cite{ghose2019mobile} conducted a large field experiment in a shopping mall to understand the effectiveness of the redeem campaign, and study the customers' trajectory within the mall.

\subsection{Location Base Advertisement and Commercial Area}

Recently, online advertisements have used location data, including the offline trajectory of the targeted customers. This capability makes online advertisements different from traditional advertisements in several aspects ~\cite{goldfarb2014different}.
Location information from mobile devices has been the source of most information of the customers’ trajectory. Recently, the importance of customers' location information for online advertisements has garnered significant attention~\cite{johnson2016location,agarwal2017geography,agarwal2011location}. This information is also effective for marketing strategies, such as coupons~\cite{molitor2020effectiveness}. In addition, the most recent papers in this topic link the location advertisement to the causal inference machine learning to assess the effect of given advertisements~\cite{kawanaka2019,moriwaki2020unbiased}.

To distribute the location-based advertisement effectively, a typical strategy sets a commercial area for potential customers because persons far from the target shop are unlikely to visit that shop. Traditionally, marketing researchers have been interested in market areas ~\cite{greenhut1952size}, and they study the evolution of the marketing area~\cite{thelen1997what} expansion~\cite{allaway1994evolution} and their driving force ~\cite{allaway2003spatial}.

Overall, previous studies found the customers' trajectory differences according to the advertisements they received. Additionally, the importance of location information for online advertisements has been indicated.  In particular, marketing scientists have mainly studied the first point with field experiments, and computer scientists have focused on the second point in the context of the ad technology. Our study aims to connect these two dots using A/B testing for O2O advertisements and offline customers' trajectory data.

\section{limitation}

The A/B testings used in this study does not force customers to visit the target shops, but the customers who are willing to do so show up in our trajectory data as we could do in some field experiments. Hence, there is a possibility that the advertisement does not alter the customers' trajectory and just picks up those who intend to change their trajectory. Therefore, it would be an oversimplification to claim that the O2O advertisements is the only source of the trajectory differences found in this study. However, as we discussed in the introduction section, the existing research continuously found that advertisements alter the behavior trajectory~\cite{hui2013the,hui2013deconstructing,ghose2019mobile}. In addition, this limitation can remain in a more rigid random control study. Even with this limitation, the body of our research manifests that the customers who received the O2O advertisements show different trajectories than the voluntary visit customers: move a wider area, travel a longer distance, and have a higher revisit ratio. We believe that our research conducts a large-scale quasi-field experiment that proposes a promising direction of behavior research in this marketing area.

\section{CONCLUSION AND DISCUSSION }

In summary, our results demonstrate that the customers who are invited by the advertisement to the physical shops show different trajectories than those who visit the shop voluntarily. In our analysis, the invited customers demonstrate a more active shopping trajectory on their visit day, whereas the voluntary customers' trajectories concentrate near the target shop. The invited customers move not just for longer distances, but also in larger areas than voluntary visit customers. These trajectory differences last even after the target shop visit day; their travel distances are approximately 6.8\% longer than the voluntary visit customers. In addition, we found that the advertisements pick up customers who have higher revisit odds, implying that the invited customers are valuable for businesses. Based on this finding, we proposed a causal inference prediction model to devise an advertisement strategy for lifting the revisiting probability. Our model does not just predict the revisit probability for a given customer, rather than how much the advertisement will increase the number of revisits, enabling us to deliver the advertisements efficiently.

These insights provide opportunities for retailers who have their physical shops and the owners of huge commercial facilities such as shopping malls. For example, the advertisements can invite new customers to their places, and such customers travel to more extensive areas that contribute to the business administrated by shopping malls. The customers who travel to more extensive areas would be exposed to more shopping spots, resulting in a high likelihood of committing unplanned purchases. These implications are also useful for the advertising industry. The price of advertisements conventionally relies on simple metrics of volume of the audiences and the resulting sales. Typical pricing of display advertisement depends the number of clicks a given ad earns (cost-per-click, CPC) or the number of conversions they earn (cost-per-action, CPA). However, our findings suggest some arbitrage between the values that advertisements bring and their actual price. For example, we could augment the value of O2O advertisements, demonstrating the economic value that would be brought to their marketplace. 

The results here are not without limitations as discussed in the previous section, calling calls for further studies. First, while we found that O2O advertisements invited customers with a more massive externality, we cannot conclude that the advertisement itself is only the cause of such externality. Second because we have no access to the customers' spending history, we are not able to estimate the economic value of the externalizes. Therefore, future work will need to collect customer spending information to estimate the economic value of invited customers to the area near the target shops. As we proposed the revisit uplifting model, one possible direction for further work is to propose a causal machine learning model that maximizes the causal externality effect by O2O advertisements.

\clearpage

\section{Appendix}
\subsection{Normalization}\label{sec:norm}

We set $y_{i}$ as 1 when the customers in treatment group visit a block $\left(u_{i}, v_{i}\right)$ more frequent than the people in the control group. To be fair, we standardize the number of visits taking considering the differences of the observed activities between Treatment and Control groups. We calculate the following value $q$ for each user. $q_{i}^{a}$, for each group $a \in \{Treatment, Control\}$each block $\left(u_{i}, v_{i}\right)$.

\begin{equation}
q_{i}^{a} =\sum_{j \in V_{i}^{a}} r^{j},
\end{equation}

where $V_{i}^{a}$ is the set of customer $j$ who are in group $a$. $r_{j}$ is the value of visit of customer $j$ calculated as 

\begin{equation}
r_{j} = \frac{1}{n_{j} \sum_{i\in C^{a}} n_{i}},
\end{equation}

where $n_j$ is the number of observation of the customer $j$ movement within the area of interests; $C^{a}$ is the set of customers in group $a$.

\subsection{The 10 Features by the SHAP Value for the Uplift Modeling}
\label{app:shap_features}
\begin{figure}[H]
    \centering
    \includegraphics[width=0.5\columnwidth, height=0.5\linewidth]{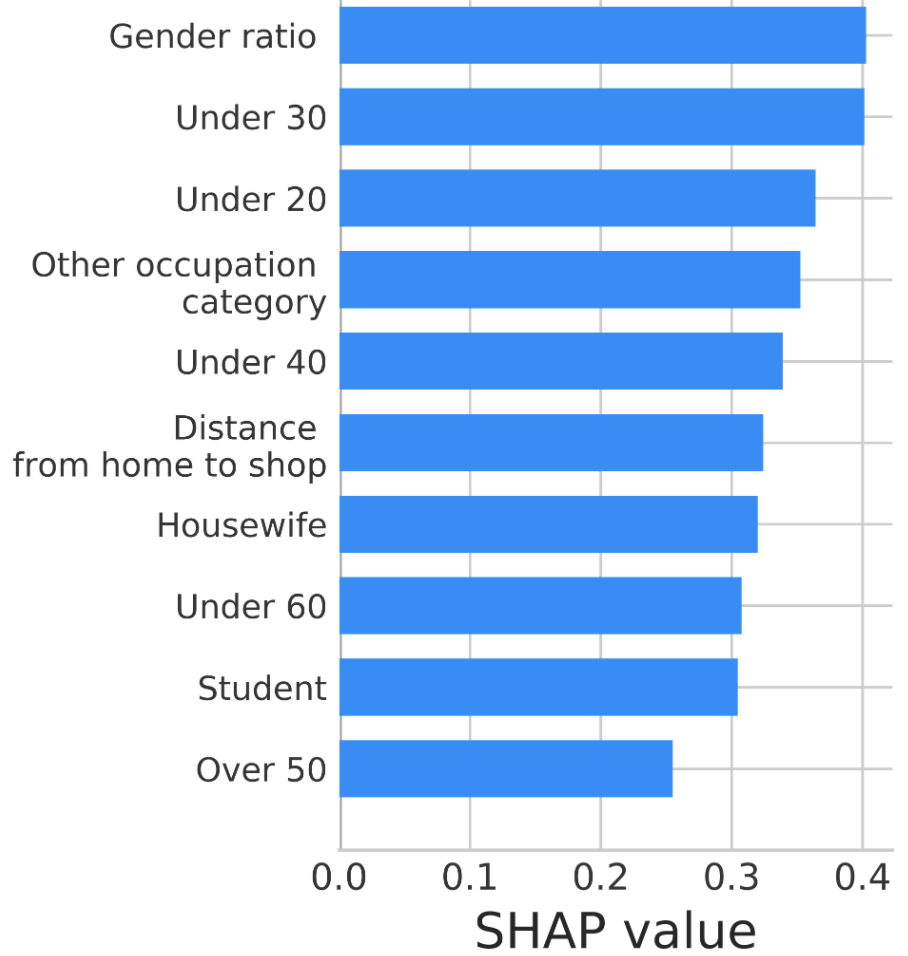}
\caption{The top 10 mean absolute SHAP value features}\label{fig:shap_mean}
\par 
\begin{minipage}{\columnwidth} %
\vskip 0.03in
{\footnotesize {\it Note:}
The top 10 mean absolute SHAP value features~\cite{shap2017} for the uplifting model. The x axis represent the absolute mean SHAP value that stands for the importance of the features for the uplift model.
\par}
\end{minipage}
\end{figure}

In this subsection, we will provide the details of the top 10 features by SHAP value~\cite{shap2017}. Table~\ref{table:desc_shap} reports the descriptions of the top 10 features in Figure~\ref{fig:sharp}. These features are the estimated demographic information and the distance from the user's home to the target shops introduced in Sec.~\ref{sec:upliftmodel_feature}. Except for ``Distance from home to shop'', the all features the estimated probability that the user has that demographic information such as being a student. We also present the mean values of the absolute SHAP values at~\ref{fig:shap_mean}

\begin{table}[H]\footnotesize
\caption{Description of the features for the uplift model}
\label{table:desc_shap} 
  \centering
    \begin{tabularx}{\columnwidth}{lX}
    \textbf{Feature} & \textbf{Description} \\
    \toprule
    Gender ratio & Estimated gender ratio of the user's living spot, \#female/(\#male + \#female)\\
    Under 20 (30, 40, 60) & Estimated probability that the user is under 20 (30, 40, 60).\\
    Over 50 & Estimated probability that the user is over 50. \\
    Other occupation category & Estimated probability that the user' occupation is not in the typical occupation category \\
    Distance from home to shop &  Estimated estimated distance from the user's home to the target shop.\\  
    Housewife &  Estimated probability that the user is housewife.\\
    Student & Estimated probability that the user is a student.\\
   \bottomrule
    \end{tabularx}
\end{table}

\end{document}